\documentclass[11pt]{asiresearch}
\usepackage[T1]{fontenc}
\usepackage{lmodern}

\usepackage{fullpage}
\usepackage{wrapfig}
\usepackage{enumitem}
\setlist[itemize]{leftmargin=*,topsep=2pt,itemsep=1pt}
\usepackage{xparse}
\RenewDocumentCommand{\paragraph}{s o m}{%
  \par\medskip\noindent\textbf{#3}\quad\ignorespaces
}

\usepackage{amsmath,amssymb}
\usepackage{chngcntr}
\usepackage{appendix}
\usepackage{titletoc}

\usepackage{tikz}
\usetikzlibrary{positioning, arrows.meta, fit, backgrounds, calc}

\definecolor{appleBlue}{HTML}{007AFF}
\definecolor{appleOrange}{HTML}{FF9500}
\definecolor{appleBg}{HTML}{F5F5F7}
\definecolor{appleText}{HTML}{1D1D1F}
\definecolor{appleBorder}{HTML}{D2D2D7}

\usepackage{algorithm}
\usepackage{algpseudocode}

\ifdefined\final
  \usepackage[disable]{todonotes}
\else
  \usepackage[textsize=tiny]{todonotes}
\fi

\newcommand{\projectpage}[1]{%
  \begin{center}
    \small
    \vspace{-1.25ex}
    \texttt{Project Page}: \url{#1}
  \end{center}
}

\title{\huge \bfseries \sffamily Residual Stream Duality in Modern \\Transformer Architectures}
\author{\textbf{Yifan Zhang}\\[1.5mm]\texttt{yifanzhangresearch@gmail.com}}
\date{}

\begin{document}
\maketitle

\begin{abstract}
Recent work has made clear that the residual pathway is not mere optimization plumbing; it is part of the model's representational machinery. We agree, but argue that the cleanest way to organize this design space is through a two-axis view of the Transformer. A decoder evolves information along two ordered dimensions: sequence position and layer depth. Self-attention already provides adaptive mixing along the sequence axis, whereas the residual stream usually performs fixed addition along the depth axis. If we fix a token position and treat layer index as the ordered variable, then a causal depth-wise residual attention read is the same local operator as causal short sliding-window attention (ShortSWA), except written over depth rather than over sequence. This is the core residual-stream duality underlying Transformer$^2$.

This perspective also clarifies the recent literature. ELC-BERT and DenseFormer already show that learned aggregation over depth can outperform uniform residual accumulation. In contrast, Vertical Attention, DeepCrossAttention (DCA), MUDDFormer, and Attention Residuals move further toward explicit attention-based routing over earlier layers. The key point, however, is that operator-level duality does not imply systems-level symmetry. For large-scale autoregressive models, sequence-axis ShortSWA is usually the more hardware-friendly placement because it reuses token-side sliding-window kernels, KV-cache layouts, and chunked execution. If the goal is instead to change the shortcut itself, Deep Delta Learning (DDL) is the cleaner intervention because it modifies the residual operator directly rather than adding a separate cross-layer retrieval path. Our recommendation is therefore simple: use DDL when the shortcut is the object of interest, and use sequence-axis ShortSWA when the goal is local adaptive mixing.
\end{abstract}

\projectpage{https://github.com/yifanzhang-pro/residual-stream-duality}

\section{Introduction}

A modern Transformer evolves information along two ordered axes: sequence position and layer depth. Along the sequence axis, self-attention performs learned, content-dependent mixing. Along the depth axis, the residual stream usually performs uniform addition. The title Transformer$^2$ is meant literally: modern Transformer architectures have two ordered directions of information flow, but only one of them is usually equipped with an adaptive attention operator. The main theme of this note is that this asymmetry is conceptually revealing and practically consequential.

That asymmetry has already motivated a broad family of proposals that replace or augment uniform depth aggregation. Earlier examples include ELC-BERT~\citep{charpentier2023not}, which feeds each layer a convex combination of earlier layer outputs, and DenseFormer, which inserts a depth-weighted average after each block~\citep{pagliardini2024denseformer}. More recent work makes the cross-depth routing explicitly attention-based, including Vertical Attention, DeepCrossAttention (DCA), MUDDFormer, and Attention Residuals \citep{kojimavertical, heddes2025deepcrossattention, xiao2025muddformer, attnres2026}. Related interventions such as Hyper-Connections and Deep Delta Learning (DDL) further underscore that shortcut design remains an active architectural degree of freedom \citep{zhu2024hyper, zhang2026deep}. The shared lesson is that the residual pathway participates in representation, not merely optimization.

\begin{figure}[t]
    \centering
    \begin{tikzpicture}[
        >=stealth,
        font=\sffamily,
        tokenNode/.style={
            circle, draw=appleBorder, fill=appleBg, thick,
            minimum size=1.1cm, inner sep=0pt, text=appleText
        },
        targetNode/.style={
            circle, draw=appleBlue, fill=appleBlue!10, very thick,
            minimum size=1.2cm, inner sep=0pt, text=appleBlue, font=\sffamily\bfseries
        },
        seqArrow/.style={->, thick, appleBlue, shorten >=2pt, shorten <=2pt},
        depthArrow/.style={->, thick, appleOrange, shorten >=2pt, shorten <=2pt},
        seqBox/.style={rounded corners=12pt, fill=appleBlue!5, draw=appleBlue!30, thick, dashed, inner sep=12pt},
        depthBox/.style={rounded corners=12pt, fill=appleOrange!5, draw=appleOrange!30, thick, dashed, inner sep=12pt}
    ]

    \draw[->, thick, appleBorder] (-1.5, -1.5) -- (8, -1.5) node[right, text=appleText, font=\sffamily\small] {Sequence Position ($t$)};
    \draw[->, thick, appleBorder] (-1.5, -1.5) -- (-1.5, 8) node[above, text=appleText, font=\sffamily\small] {Layer Depth ($\ell$)};

    \node[tokenNode] (h00) at (0,0) {$h_{t-2}^{(\ell-2)}$};
    \node[tokenNode] (h01) at (3,0) {$h_{t-1}^{(\ell-2)}$};
    \node[tokenNode] (h02) at (6,0) {$h_{t}^{(\ell-2)}$};

    \node[tokenNode] (h10) at (0,3) {$h_{t-2}^{(\ell-1)}$};
    \node[tokenNode] (h11) at (3,3) {$h_{t-1}^{(\ell-1)}$};
    \node[tokenNode] (h12) at (6,3) {$h_{t}^{(\ell-1)}$};

    \node[tokenNode] (h20) at (0,6) {$h_{t-2}^{(\ell)}$};
    \node[tokenNode] (h21) at (3,6) {$h_{t-1}^{(\ell)}$};
    \node[targetNode] (h22) at (6,6) {$h_{t}^{(\ell)}$};

    \begin{scope}[on background layer]
        \node[seqBox, fit=(h20) (h21) (h22)] (sBox) {};
        \node[above=2mm of sBox.north, text=appleBlue, font=\sffamily\bfseries\small] {Sequence Axis: Adaptive Mixing (Attention)};

        \node[depthBox, fit=(h02) (h12) (h22)] (dBox) {};
        \node[right=9mm of dBox.east, text=appleOrange, font=\sffamily\bfseries\small, rotate=-90, anchor=north] {Depth Axis: Residual Attention (ShortSWA)};
    \end{scope}

    \draw[seqArrow, bend left=25] (h20.north) to (h22.north);
    \draw[seqArrow] (h21.east) to (h22.west);

    \draw[depthArrow, bend right=25] (h02.east) to (h22.east);
    \draw[depthArrow] (h12.north) to (h22.south);

    \node[align=center, text=appleText, font=\sffamily\small, text width=5cm] at (2, 1.5) {
        \textbf{Transformer$^2$ Duality}\\
        The local operator is identical.\\
    };

    \end{tikzpicture}
    \caption{\textbf{Overview of Residual Stream Duality.} A modern Transformer evolves information along two ordered dimensions: sequence position and layer depth. Computing the target state $h_t^{(\ell)}$ via explicit depth-wise residual attention operates identically to causal short sliding-window attention (ShortSWA) on the sequence axis, effectively bridging the representation mechanisms of the two pathways.}
    \label{fig:overview_duality}
\end{figure}
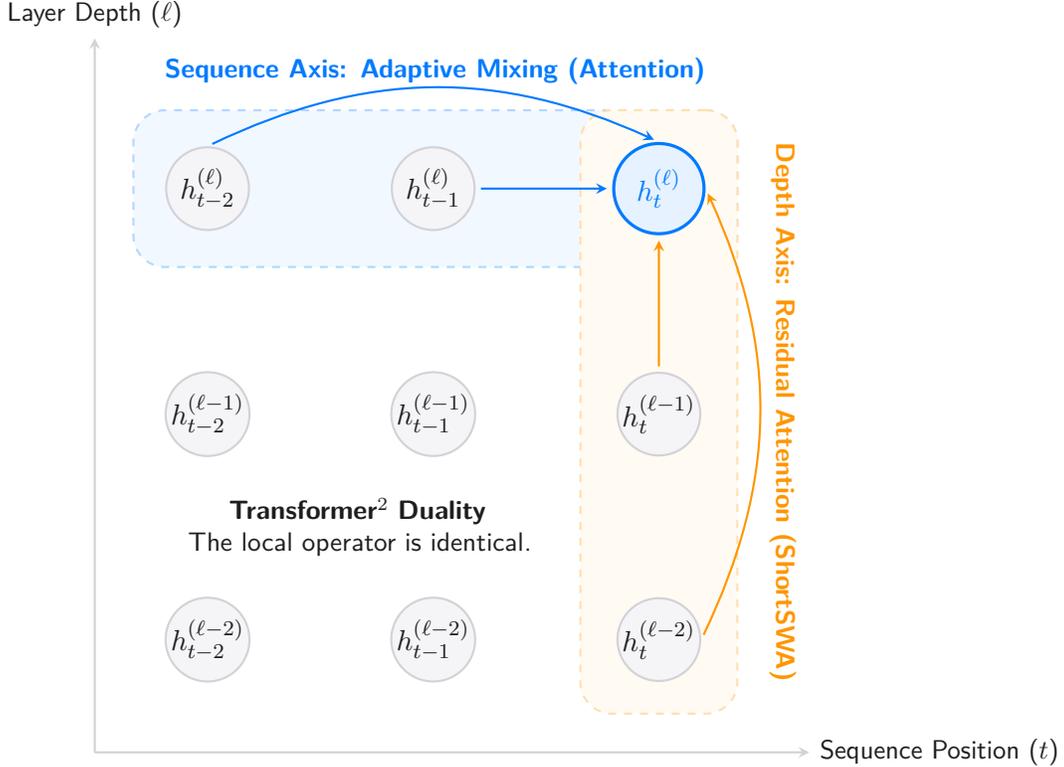

Our claim is not that all of these proposals are identical. ELC-BERT and DenseFormer sit on the learned-static end of the spectrum; Vertical Attention, DCA, MUDDFormer, and Attention Residuals use more expressive routing modules. But the common object is learned aggregation over the ordered depth axis. The cleanest exact statement applies to explicit depth-wise attention reads: once a token position is fixed, and a layer index is treated as a one-dimensional ordered axis, a truncated residual attention read is precisely causal ShortSWA written over depth. The full-memory variant is simply the full-window limit of the same operator family.

That duality is mathematical, not systems-symmetric. Sequence-axis ShortSWA reuses existing sliding-window attention kernels, token-side KV-cache layouts, and chunked execution strategies. Depth-axis aggregation, by contrast, requires an additional layer-indexed state path: each block needs online access to earlier layer states or block summaries for the same token, and under pipeline parallelism, those states may need to be forwarded, stored, or recomputed. The practical question is therefore not whether attention can be applied over depth, but whether depth is the right axis on which to place a short adaptive mixer. The thesis of this note is therefore:
\begin{itemize}
    \item A depth-wise residual attention read is not a new local operator. It is ShortSWA written on the depth axis rather than the sequence axis.
    \item Learned cross-depth aggregation spans a continuum from static depth weighting (ELC-BERT, DenseFormer) to attention-based routing (Vertical Attention, DCA, MUDDFormer, Attention Residuals). These systems are not identical end-to-end, but they occupy the same design space.
    \item Once that distinction is explicit, the natural design choice is either to use Deep Delta Learning~\citep{zhang2026deep} to improve the shortcut itself or to place ShortSWA directly on the sequence axis, which is usually more hardware-efficient for current training and inference stacks.
    \item Following \citet{zhang2025rethinking}, we view ShortSWA as the successor to ShortConv and, in spirit, the attention-form successor to Canon layers \citep{allen2025physics}.
\end{itemize}

\paragraph{Relation to prior depth-aggregation work.}
ELC-BERT and DenseFormer are important precursors because they already replace uniform depth accumulation with learned aggregation. ELC-BERT feeds each layer a convex combination of previous layer outputs, while DenseFormer adds a depth-weighted average of current and past representations after each block~\citep{charpentier2023not, pagliardini2024denseformer}. Vertical Attention, DCA, MUDDFormer, and Attention Residuals move further toward attention-based routing over earlier layers~\citep{kojimavertical, heddes2025deepcrossattention, xiao2025muddformer, attnres2026}. Our claim is therefore not that these methods are end-to-end identical. It is that, once depth is treated as an ordered axis, they are best compared inside one common design space of learned cross-depth aggregation. DDL, by contrast, attacks a different target: it changes the shortcut update itself rather than adding a separate retrieval path over stored earlier states~\citep{zhang2026deep}. Hyper-Connections make a related point, that residual design is itself a meaningful architectural degree of freedom, but they do not remove the system's asymmetry between token-side local mixing and layer-side state management~\citep{zhu2024hyper}.

\paragraph{Relation to ShortConv and Canon layers.}
ShortConv, Canon layers, and ShortSWA all occupy the same architectural slot: they are local mixers that operate before or alongside a broader global mechanism. ShortConv uses a fixed, small kernel. Canon layers compute learned weighted sums over nearby tokens \citep{allen2025physics}. As argued in \citet{zhang2025rethinking}, once chunked computation is already part of the implementation, the natural upgrade is ShortSWA: the same local role, but with content-adaptive mixing and a chunk-aligned receptive field. In that sense, ShortSWA is the natural successor to ShortConv and the attention-form successor to Canon layers.

\section{Residual Stream Duality}

\subsection{Preliminaries}

Let $H = [H^{(0)}, \ldots, H^{(L)}] \in \mathbb{R}^{(L+1) \times T \times d}$ denote the hidden-state stack of an $L$-block decoder, where $H^{(0)}$ is the input stream, $T$ is the sequence length, and $d$ is the model width. We write $H^{(\ell)} \in \mathbb{R}^{T \times d}$ for the hidden states at depth $\ell$. A standard pre-norm Transformer block, for $\ell = 0, \ldots, L-1$, is
\[
U^{(\ell)} = H^{(\ell)} + \mathrm{Attn}\!\left(\mathrm{Norm}\!\left(H^{(\ell)}\right)\right),
\]
\[
H^{(\ell+1)} = U^{(\ell)} + \mathrm{MLP}\!\left(\mathrm{Norm}\!\left(U^{(\ell)}\right)\right).
\]
The sequence axis is mixed adaptively by attention, while the depth axis is mixed by fixed addition.

\subsection{Depth-wise residual attention is ShortSWA on the depth axis}

Fix a token position $t$ and collect its trajectory through depth:
\[
X_t = [h_t^{(0)}; h_t^{(1)}; \cdots; h_t^{(L)}] \in \mathbb{R}^{(L+1) \times d}.
\]
Now consider a causal depth window of size $K$. Define
\[
\mathcal{W}_{t,\ell}^{(K)} =
[h_t^{(\max(0,\ell-K+1))}; \ldots; h_t^{(\ell)}]
\in \mathbb{R}^{K_\ell \times d},
\qquad
K_\ell = \min(K,\ell+1).
\]
A depth-wise residual attention read at layer $\ell$ can be written as
\[
q_{t,\ell} = W_Q h_t^{(\ell)}, \qquad
\mathbf{K}_{t,\ell} = \mathcal{W}_{t,\ell}^{(K)} W_K, \qquad
\mathbf{V}_{t,\ell} = \mathcal{W}_{t,\ell}^{(K)} W_V,
\]
\[
z_t^{(\ell)} =
\mathrm{softmax}\!\left(
\frac{q_{t,\ell} \mathbf{K}_{t,\ell}^{\top}}{\sqrt{d_k}}
\right)\mathbf{V}_{t,\ell}.
\]
This is exactly causal ShortSWA applied to the one-dimensional sequence $X_t$ whose index is the layer number:
\[
z_t^{(\ell)} = \mathrm{ShortSWA}\!\bigl(X_t; K\bigr)_\ell.
\]
Hence, after transposing the hidden-state tensor so that depth becomes the ordered axis, truncated depth-wise residual attention and ShortSWA belong to the same operator family. The full-memory residual attention variant is simply the full-window limit $K=\ell+1$.

This exact equivalence applies whenever the cross-depth retrieval is implemented as an explicit attention read. Simpler learned-weight schemes such as ELC-BERT or DenseFormer belong to the same broader design space but are not literally instances of the QKV operator above \citep{charpentier2023not, pagliardini2024denseformer}.

\subsection{A unified view of learned depth aggregation}

The exact equivalence above applies to the explicit depth-wise residual attention read written here. It also suggests a useful taxonomy of nearby methods. ELC-BERT and DenseFormer are learned depth aggregators with parameterized weights over earlier layers, but without a full depth-wise QK attention read \citep{charpentier2023not, pagliardini2024denseformer}. Vertical Attention, DCA, MUDDFormer, and Attention Residuals are closer to the explicit attention end of the spectrum: Vertical Attention learns inter-layer paths through routing modules, DCA computes attention inputs from mixtures of previous layer outputs, MUDDFormer introduces separate dynamic dense modules for query, key, value, and residual streams, and Attention Residuals presents the read most directly as attention over depth \citep{kojimavertical, heddes2025deepcrossattention, xiao2025muddformer, attnres2026}. These systems are not identical end-to-end architectures; they differ in factorization, parameter sharing, gating, and injection point. What the duality statement contributes is a common coordinate system: once depth is treated as an ordered axis, explicit cross-depth attention is simply local causal attention on that axis, and the broader family can be read as increasingly expressive parameterizations of learned depth aggregation.

\subsection{Why the sequence axis is the better placement}

Once the equivalence above is explicit, the main design question becomes where to place the short attention primitive when the goal is local adaptive mixing. Our view is that the sequence axis is the better answer:
\[
S^{(\ell)} = H^{(\ell)} +
\mathrm{ShortSWA}\!\bigl(\mathrm{Norm}(H^{(\ell)}); w\bigr),
\]
\[
A^{(\ell)} = S^{(\ell)} + \mathrm{Attn}\!\left(\mathrm{Norm}\!\left(S^{(\ell)}\right)\right),
\]
\[
H^{(\ell+1)} = A^{(\ell)} + \mathrm{MLP}\!\left(\mathrm{Norm}\!\left(A^{(\ell)}\right)\right).
\]
This preserves the same local-to-global story but places the adaptive local mixer on the axis that modern kernels and inference stacks already optimize.

At autoregressive inference time, sequence-axis ShortSWA can reuse the usual token-side cache layout over the most recent $w$ tokens. In chunked training or inference, the local window can be aligned to the chunk already loaded into SRAM. Under pipeline parallelism, the implementation preserves the standard forward flow of activations between layer partitions rather than introducing an additional layer-indexed state path. Depth-axis attention-style aggregation faces the opposite incentives: each block needs online access to earlier layer states or block summaries for the same token. Methods such as Vertical Attention, DCA, MUDDFormer, and blockwise Attention Residuals differ in how they parameterize or compress this access, but they all live with the same underlying pressure: depth-side routing must manage cross-layer state explicitly \citep{kojimavertical, heddes2025deepcrossattention, xiao2025muddformer, attnres2026}. If the target is instead the shortcut operator itself, we would choose Deep Delta Learning rather than add another cross-depth read, because DDL changes the residual update directly and does not require an explicit stack of earlier layer states \citep{zhang2026deep}.

\subsection{Recommended block}

The resulting recommendation is therefore a clean two-way design fork:
\begin{itemize}
    \item If the goal is a better shortcut, use Deep Delta Learning \citep{zhang2026deep}.
    \item If the goal is a local content-adaptive mixer, use ShortSWA directly on the sequence axis.
\end{itemize}
For current large-scale training and inference stacks, we do not see a general systems case for a third option that repackages sequence-local attention as a depth-axis residual mechanism. When the goal is sequence-side local mixing, the resulting local-to-global Transformer block is as follows.

\begin{algorithm}[ht!]
\caption{Recommended local-to-global block when the goal is sequence-side local mixing}
\begin{algorithmic}[1]
\Require hidden states $H^{(\ell)}$, local window $w$
\State $S^{(\ell)} \gets H^{(\ell)} + \mathrm{ShortSWA}(\mathrm{Norm}(H^{(\ell)}); w)$
\State $A^{(\ell)} \gets S^{(\ell)} + \mathrm{Attn}(\mathrm{Norm}(S^{(\ell)}))$
\State $H^{(\ell+1)} \gets A^{(\ell)} + \mathrm{MLP}(\mathrm{Norm}(A^{(\ell)}))$
\end{algorithmic}
\end{algorithm}

\subsection{Complexity and systems notes}

Ignoring head-wise constants, ShortSWA adds a local attention term of roughly $O(Twd)$ per layer, where $w \ll T$. If a block still includes full self-attention, the asymptotic sequence-mixing cost remains $O(T^2 d)$ up to constant factors. The important point is not a new asymptotic regime, but a better hardware placement: the local operation lives on the token axis and can reuse standard sliding-window kernels and KV-cache layouts.

The next estimates apply to explicit attention-style reads over depth, not to lighter learned-weight schemes such as ELC-BERT or DenseFormer. Depth-wise residual attention with a depth window $K$ adds $O(TKd)$ work per block, hence $O(TKLd)$ across an $L$-block network, together with additional online access to earlier layer states or block summaries. The full-depth variant grows to $O(TL^2 d)$ for the score/value interactions. These formulas make the compute overhead visible, but the more consequential issue in practice is systems complexity: one now needs an extra layer-indexed state that must be retained, forwarded, or recomputed, especially when depth windows cross pipeline-stage boundaries. In some deployments, this behaves like a second cache over depth. DDL avoids this depth-axis state-management overhead because it modifies the per-block shortcut rather than attending over stored earlier layer states~\citep{zhang2026deep}.

\section{Conclusion}

This note formalizes a simple residual-stream duality. A Transformer evolves information along two ordered axes: sequence position and layer depth. When a token position is fixed, and the layer index is treated as the ordered variable, an explicit depth-wise residual attention read is ShortSWA on the transposed axis: a causal local attention operator over previous layers rather than previous tokens. This view places prior depth-aggregation methods in one design space. ELC-BERT and DenseFormer use learned but mostly static depth aggregation, while Vertical Attention, DCA, MUDDFormer, and Attention Residuals move toward content-dependent cross-depth routing~\citep{charpentier2023not, pagliardini2024denseformer, kojimavertical, heddes2025deepcrossattention, xiao2025muddformer, attnres2026}. They differ architecturally, but all modify how information is aggregated across depth.

The duality is therefore conceptual, not a mandate to attend over depth by default. If the goal is to improve the shortcut itself, DDL is the more direct intervention. If the goal is local adaptive mixing, sequence-axis ShortSWA is usually the cleaner systems choice, since it matches existing sliding-window kernels, token-side KV caches, and chunked execution. In this sense, ShortSWA remains the natural successor to ShortConv and the content-adaptive analogue of Canon-style local mixing \citep{zhang2025rethinking, allen2025physics}. Transformer$^2$ is best read as an organizing principle: one local attention family, two ordered axes, and a practical preference for sequence placement unless cross-depth retrieval is the object of study.

\section*{Acknowledgement}
We sincerely thank Xinyu Yang for helpful discussions. We used large language models to assist in polishing the writing of this work.

\vspace{5ex}
\bibliographystyle{plainnat}
\bibliography{reference}

\clearpage
\appendix

\renewcommand{\appendixpagename}{\centering \huge Appendix}
\appendixpage
\counterwithin{theorem}{section} 

\startcontents[section]
\printcontents[section]{l}{1}{\setcounter{tocdepth}{2}}
\clearpage

\end{document}